\pdfoutput=1

\documentclass[11pt]{article}

\usepackage[]{eacl2023}

\usepackage{times}
\usepackage{latexsym}
\usepackage{arydshln}
\usepackage{graphicx}
\usepackage[T1]{fontenc}
\usepackage{array}
\usepackage{verbatim}

\usepackage[utf8]{inputenc}
\usepackage{multirow}
\usepackage{multicol}

\usepackage{microtype}

\usepackage{inconsolata}
\usepackage{amsmath}
\usepackage{amssymb}
\usepackage{booktabs}
\usepackage{graphicx}
\usepackage{amsmath}
\usepackage{tabularx}
\usepackage{hyperref}
\usepackage{float}
\usepackage[normalem]{ulem}
\useunder{\uline}{\ul}{}
\usepackage{graphics}

\title{CoTEVer: Chain of Thought Prompting Annotation Toolkit \\for Explanation Verification}

\author{Seungone Kim$^{1,2}$ \quad Sejune Joo$^{1,2}$ \quad Yul Jang$^{2}$ \quad Hyungjoo Chae$^{2}$ \quad Jinyoung Yeo$^{2}$ \\ \\ KAIST AI$^{1}$ \quad Yonsei University$^{2}$\\
  \texttt{louisdebroglie@kaist.ac.kr}\\
  \texttt{\{sr7418,blaze,mapoout,jinyeo\}@yonsei.ac.kr}\\}
  
\begin{document}
\maketitle

\begin{abstract}
%



Chain-of-thought (CoT) prompting enables large language models (LLMs) to solve complex reasoning tasks by generating an explanation before the final prediction. Despite it's promising ability, a critical downside of CoT prompting is that the performance is greatly affected by the factuality of the generated explanation. To improve the correctness of the explanations, fine-tuning language models with explanation data is needed. However, there exists only a few datasets that can be used for such approaches, and no data collection tool for building them. Thus, we introduce \textbf{CoTEVer}, a tool-kit for annotating the factual correctness of generated explanations and collecting revision data of wrong explanations. Furthermore, we suggest several use cases where the data collected with \textbf{CoTEVer} can be utilized for enhancing the faithfulness of explanations. Our toolkit is publicly available at~\href{https://github.com/SeungoneKim/CoTEVer}{https://github.com/SeungoneKim/CoTEVer}.
\end{abstract}

\section{Introduction}

Chain-of-thought prompting~\citep{wei2022chain} generates an explanation before the answer to elicit the reasoning capabilities of large language models. An intuitive way to interpret chain-of-thought prompting is that the process of `\textit{generating an explanation}' is analogous to `\textit{decomposing multiple step problems into smaller sub-problems}', which enables to solve complex reasoning tasks. Therefore, generating a plausible explanation is crucial to derive the correct answer~\citep{wang2022self}.

To generate a plausible explanation, previous works have attempted to generate multiple explanations and use a task-specific verifier that would access the quality of the explanations and choose one of them~\citep{cobbe2021training,shen2021generate,thoppilan2022lamda,li2022advance}. A more fundamental solution to this problem is fine-tuning the underlying language model with high-quality annotated explanations~\citep{ling2017program, cobbe2021training, zelikman2022star, huang2022large, chung2022scaling}. However, fine-tuning would require to gather large amounts of annotated explanation data, which is impractical.


\begin{figure}[t!]
\centering
    \includegraphics[width=0.95\linewidth]{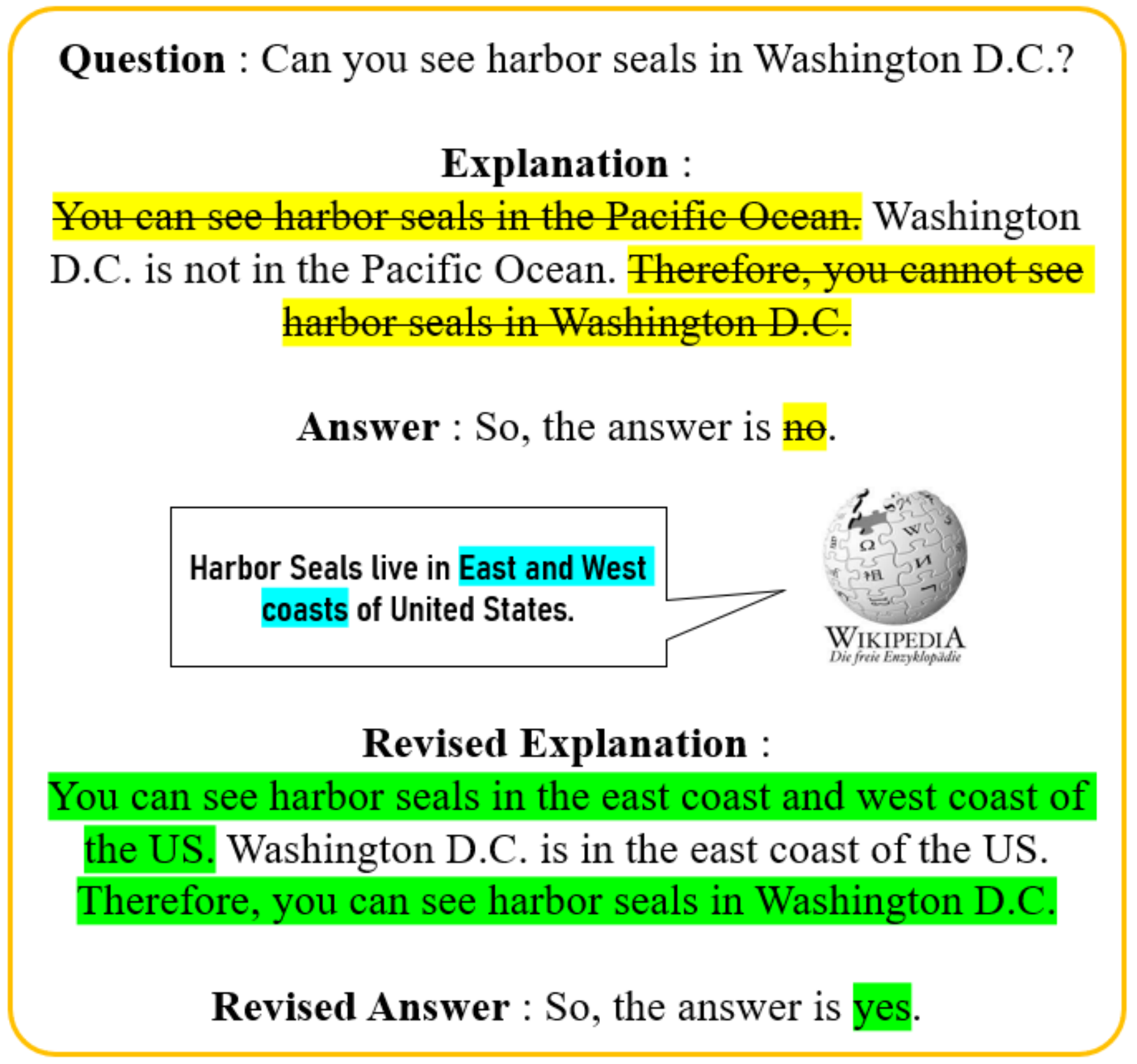}
    \caption{Example of Explanation Verification and Answer Verification of GPT-3's output. Explanation Verification requires additional knowledge which makes it hard for annotators to intuitively write a revised explanation and answer.}
    \label{fig:example}
\end{figure}

Collecting large amounts of annotated explanation data is difficult for several reasons. First, while existing works gather explanation data by asking annotators to manually write explanations using existing datasets~\citep{wiegreffe2021teach}, gathering human authored labels is often expensive in terms of time and cost~\citep{west2021symbolic}. Second, writing a good quality explanation from scratch is difficult because it requires sufficient background knowledge~\citep{geva2021did}. 

In this paper, we address the question: can we gather explanation data in a more \textit{efficient} manner? Inspired by human-in-the-loop methods, we ask annotators to verify a machine generated explanation instead of manually writing them~\citep{wallace2019trick, weber2021better, du2022read}. In other words, annotators get to check whether the underlying language model \textit{hallucinate} (i.e., generate explanations that are factually incorrect)~\citep{shuster2021retrieval,lin2022truthfulqa}. To do this, we provide a set of supporting evidence documents retrieved from the web. Annotators access the quality of the given explanation, and provide a feedback score along with a better alternative.


As shown in Figure~\ref{fig:example}, let's consider gathering an explanation and answer for the question, `Can you see harbor seals in Washington D.C.?'\footnote{Example from StrategyQA~\citep{geva2021did}}. In this example, GPT-3 generates an explanation `1) You can see harbor seals in the Pacific Ocean. 2) Washington D.C. is not in the Pacific Ocean. 3) Therefore you cannot see harbor seals in Washington D.C.' and predicts `No' as the answer. In this case, the first sentence of the explanation missed the point that harbor seals not only live in the west coast, but also in the east coast of the US. By providing the background knowledge `Harbor Seals live in east and west coasts of United States', annotators could successfully revise the explanation.

To this end, we propose \textbf{CoTEVer} (\underline{C}hain \underline{o}f \underline{T}hought Prompting Annotation Toolkit for \underline{E}xplanation \underline{Ver}ification), which is designed to \textit{efficiently} gather explanation data, by 1) alleviating the role of annotators to verify instead of writing from scratch and 2) supplementing the required background knowledge via evidence documents. With the  gathered explanation data, researchers could use them for CoT fine-tuning~\citep{chung2022scaling} or transform them into other knowledge intensive datasets.



\section{Related Works}

\subsection{Tool-kits for Data Annotation}

There exists a number of interactive tool-kits for annotating and verifying labels~\citep{gotze2022slurk,lin2022dotat,friedrich2021annie,bach2022promptsource,thrush2022dynatask}. For instance, Promptsource~\citep{bach2022promptsource}, is a framework designed to try out diverse set of prompts that can be used in in-context learning~\citep{liu2021makes}, or instruction tuning~\citep{sanh2021multitask,wei2021finetuned,min2021metaicl,ye2022guess,jang2023exploring}.  Other human-in-the-loop annotation toolkits~\citep{wallace2019trick, weber2021better, du2022read} provides functionality for annotators to verify the neural model's prediction instead of manually creating them. Compared to these toolkits, \textbf{CoTEver} provides additional features specifically designed for gathering explanation data such as retrieving evidence documents and supporting different Chain of Thought prompts.

\subsection{Explanation Data}

Chain of Thought Prompting is an in-context learning based methodology that generates an explanation before the answer. Instead of directly answering to the question, \citet{wei2022chain} conjectures that generating an explanation on-the-fly (explain-and-generate) enhances the reasoning capabilities of large language models. \citet{wei2022emergent} argues that the ability to solve complex reasoning only appears when using large-scale language models, and defines this phenomenon as `\textit{Emergent Abilities}'. \textbf{CoTEver} uses Chain of Thought Prompting to generate an explanation that could serve as a starting point for annotators to verify.


Recently, \citet{chung2022scaling} has shown that fine-tuning with explanation data unlocks the emergent abilities in large language models and achieves good performance not only at seen tasks~\citep{ling2017program, cobbe2021training, zelikman2022star}, but also unseen tasks. The explanation data collected by \textbf{CoTEVer} could be used for CoT Finetuning since we collect a revised explanation.



\subsection{Hallucination in Language Models}

Hallucination is a phenomenon where a model generates a falsehood output that may contradict with the factual knowledge. \citet{lin2022truthfulqa} reported that as the model size increases, the less truthful they tend to be. \citet{lewis2020retrieval} explains that models that rely only on parametric memory (e.g., GPT-3) are more likely to suffer from hallucination. When collecting explanation data from annotators, hallucination is a critical issue because the model may generate an unfaithful but very fluent output that is not easily distinguishable~\citep{gao2022attributed}. To collect factually correct explanations from annotators, we provide supporting evidence documents using a search engine.

\begin{figure*}[t!]
\centering
    \includegraphics[width=0.99\linewidth]{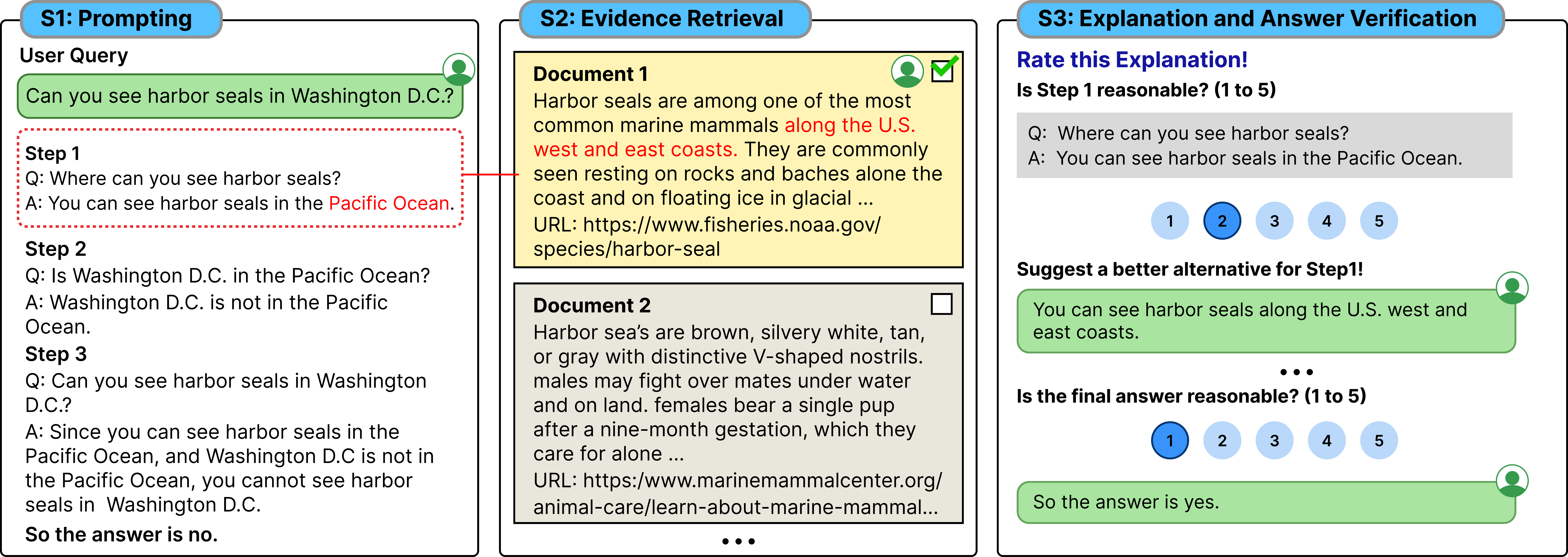}
    \caption{The overall illustration of \textbf{CoTEver}. An annotator asks a question to \textbf{CoTEver} and receives an explanation, supporting evidence documents, and a prediction. Then, the annotator's rating of the explanation (5 for most relevant), suggestions for a better explanation is stored in the Database which can be used for research purposes.}
    \label{fig:overview}
\end{figure*}
\section{System Design and Workflow}
In Figure~\ref{fig:overview}, we present an illustration of the overall explanation verification process of \textbf{CoTEver} with 3 steps and show how the annotated explanations could be obtained effectively. We assume a scenario where a researcher requests a group of annotators to query a large language model and verify the explanations and predictions to collect explanation data. Although \textbf{CoTEVer} could support gathering free-form questions from annotators, it would either require 1) the researcher to make predefined few-shot demonstrations and retrieving them on-the-fly or 2) generating the explanation in a zero-shot setting~\citep{kojima2022large}, which is both challenging to gather good quality explanations. Therefore, we define a scenario where a researcher assign users to query specific type of questions, such as `Ask a question that could be answered with yes/no'(Answer Format) or `Ask a question that is related to economics'(Domain). In this case, we could assume that the researcher prepared few-shot demonstrations beforehand.

\subsection{S1: Prompting}\label{section3.1:prompting}
\noindent{\textbf{Prompting Composition.}}
We use GPT-3~\citep{brown2020language} which is one of the standard large language models for CoT prompting~\citep{wei2022chain,kojima2022large}. CoT prompting has limitations in that the generated explanations does not have a unified format, which makes verification harder. So, we adopt Self Ask~\citep{press2022measuring} as our prompting method to generate explanations composed of sub-questions and sub-answers. We provide more details of our prompt in Table~\ref{table:strategy_qa_prompt}.\\
\noindent{\textbf{Explanation Generation.}}
As a first step, annotators are asked to explore our framework by querying a variety of different questions under the designated requirements. The user question is concatenated to the demonstrations as an input for the model. We then send a request via OpenAI API and get a response which contains the output of GPT-3. Upon obtaining the responses, we decompose the output into reasoning steps consist of a sub-question and sub-answer.

\subsection{S2: Evidence Retrieval}

\noindent{\textbf{Document Retrieval.}} To help the annotators' judgement, we provide documents that can be used as evidence to the generated explanation. For that, we retrieve documents using the sub-question directly as a query for document retrieval. Note that while \citet{press2022measuring} use the sub-questions to retrieve and answer to the question, we instead use them to retrieve and help annotators verify the explanations. In order to retrieve documents that are not outdated from the Web database, we use Google Search API~\footnote{https://pypi.org/project/google-api-python-client/}. We use the top-10 documents as candidates for reranking.\\

\noindent{\textbf{Document Reranking.}} We first split the retrieved documents into smaller chunks of maximum 512 tokens, the maximum number of input tokens acceptable in our reranking module. To provide reliable evidence documents to the annotators, we rerank the documents candidates and choose the order to be displayed on the user interface. We obtain the embeddings of the query and the supporting document candidates using Sentence-T5~\citep{ni2021sentence}, a state-of-the-art sentence embedding model trained to capture the semantic textual similarity between two input sequences. We then rank the retrieved documents with a cosine similarity score between the document and query embedding.

\begin{table*}[t!]
\begin{center}\small
\setlength{\extrarowheight}{2pt}
{\begin{tabular}{|c|l|l|} 
\toprule
\textbf{Error Type} & \multicolumn{2}{c|}{\textbf{Example}} \\ 
\midrule
\multirow{11}{*}[-2.5ex]{\begin{tabular}[c]{@{}c@{}}\textbf{Insufficient }\\ \textbf{ Knowledge}\\ \textbf{18\%} \end{tabular}} & \multicolumn{2}{c|}{\textbf{Do all crustaceans live in the ocean?}} \\ 
\cline{2-3}
 & Q\#1: What are crustaceans? & \multirow{6}{*}{\begin{tabular}[c]{@{}l@{}}\textcolor{blue}{Crabs} can be found in all \textcolor{red}{oceans} \textcolor{red}{and} \textcolor{red}{in}\\ \textcolor{red}{freshwater}. Some crabs live on land,\\ sometimes several miles from water.\\ Species such as the fiddler crab live in \\ burrows in the sand or mud, where \\ they stay during winter and high tides.\end{tabular}} \\
 & \begin{tabular}[c]{@{}l@{}}A\#1: Crustaceans are a group of animals that \\ \qquad \ include crabs, lobsters, and shrimp.\end{tabular} &  \\
 & Q\#2: Where do crabs live? &  \\
 & A\#2: \textcolor{blue}{Crabs} live in the \textcolor{red}{ocean}. &  \\
 & Q\#3: Where do lobsters live? &  \\
 & A\#3: Lobsters live in the ocean. &  \\ 
\cline{3-3}
 & Q\#4: Where do shrimp live? & \multirow{2}{*}{\begin{tabular}[c]{@{}l@{}}\textcolor{blue}{Shrimp} occur in all \textcolor{red}{oceans}—in shallow \\ and deep water—\textcolor{red}{and} \textcolor{red}{in} \textcolor{red}{freshwater }\\\textcolor{red}{ lakes and streams}.\end{tabular}} \\
 & A\#4: \textcolor{blue}{Shrimp} live in the \textcolor{red}{ocean}. &  \\
 & Q\#5: Do all crustaceans live in the ocean? &  \\
 & \begin{tabular}[c]{@{}l@{}}A\#5: Since crabs, lobsters, and shrimp live in the \\ \qquad \  ocean, all crustaceans live in the ocean.\end{tabular} &  \\ 
\midrule
\multirow{7}{*}[-5ex]{\begin{tabular}[c]{@{}c@{}}\textbf{Out of Date}\\\textbf{22\%}\end{tabular}} & \multicolumn{2}{c|}{\textbf{Could a single bitcoin ever cover cost of a Volkswagen Jetta?}} \\ 
\cline{2-3}
 & Q\#1: How much is a single bitcoin? & \multirow{3}{*}{\begin{tabular}[c]{@{}l@{}}\textcolor{blue}{Bitcoin’s price} today is \textcolor{red}{\$ 21,282.48},\\ with a 24-hour trading volume of \\ \$ 37.84 B.\end{tabular}}\\
 & A\#1: \textcolor{blue}{A single bitcoin} is about \textcolor{red}{\$6,000}. &  \\
 & Q\#2: How much is a Volkswagen Jetta? &  \\
 & A\#2: A Volkswagen Jetta is about \$18,645. &  \\
 & \begin{tabular}[c]{@{}l@{}}Q\#3: Could a single bitcoin ever cover cost \\ \qquad \ of a Volkswagen Jetta?\end{tabular} &  \\
 & \begin{tabular}[c]{@{}l@{}}A\#3: Since a single bitcoin is about \$ 6,000, \\ \qquad \ and a Volkswagen Jetta is about \$ 18,645, \\ \qquad \ and \$ 6,000 is smaller than \$ 18,645, \\ \qquad \ single bitcoin could not cover the cost \\ \qquad \ of a Volkswagen Jetta.\end{tabular} &  \\ 
\midrule
\multirow{9}{*}[-7ex]{\begin{tabular}[c]{@{}c@{}}\textbf{Wrong Fact}\\\textbf{43\%}\\\end{tabular}} & \multicolumn{2}{c|}{\textbf{Can a lifeboat rescue people in the Hooke Sea?}} \\ 
\cline{2-3}
 & Q\#1: What is the Hooke Sea? & \multirow{4}{*}{\begin{tabular}[c]{@{}l@{}}\textcolor{blue}{Hooke Crater} is an \textcolor{red}{impact crater} in the \\ Argyre quadrangle \textcolor{red}{on Mars} at 45.2°S\\ and 44.4°W and is 139.0 km in diameter. \\ It was named after British physicist-\\ astronomer Robert Hooke.[1]\end{tabular}} \\
 & A\#1: \textcolor{blue}{The Hooke Sea} is a sea in \textcolor{red}{Antarctica}. &  \\
 & Q\#2: Is the Hooke Sea a dangerous place? &  \\
 & \begin{tabular}[c]{@{}l@{}}A\#2: The Hooke Sea is a dangerous place \\\qquad \  because it is a sea in Antarctica.\end{tabular} &  \\
 & Q\#3: What is a lifeboat? &  \\
 & \begin{tabular}[c]{@{}l@{}}A\#3: A lifeboat is a boat designed for \\\qquad \  rescuing people in danger.\end{tabular} &  \\
 & Q\#4: Can a lifeboat rescue people in the Hooke Sea? &  \\
 & \begin{tabular}[c]{@{}l@{}}A\#4: Since the Hooke Sea is a dangerous place \\ \qquad \ in Antarctica, and a lifeboat is a boat \\ \qquad \ designed for rescuing people in danger, \\ \qquad \ lifeboat can rescue people in \\ \qquad \ the Hooke Sea.\end{tabular} &  \\
\bottomrule
\end{tabular}}                 
\end{center}
\caption{Examples of frequent error types within explanations. The left side is the original explanation generated by a language model, and the right side is the retrieved evidence document. The evidence documents could provide the required knowledge when revising the original explanation.}
\label{tab:update}
\end{table*}

\subsection{S3: Explanation and Answer Verification}
\noindent{\textbf{Explanation verification.}} 
In light of the provided evidence documents, annotators could easily check the correctness of the given explanation and give a 1-5 star Likert scale rating for each step in the explanation. In cases where the explanation needs to be revised, annotators can submit a better alternative. Our main intuition is that compared to writing a new explanation from scratch, revising an explanation with the evidence documents is much more easier for the annotators. Also, it is more likely that the revised explanation would be factually correct because the evidence documents would ground the required knowledge during annotation. The rating data is further used to determine the quality of a given explanation, which we further explain in Section~\ref{sec:5}

During the verification process, the annotators also label which evidence document is used as an evidence. For example, in Step 2 of Figure~\ref{fig:overview}, the annotator put a check mark on the document that contains the information about the habitat of harbor seals which contradicts to the sub-answer in the first step, ``\textit{You can see harbor seals in the Pacific Ocean.}''. We further explain how this data could be utilized in Section~\ref{sec:5}.

\noindent{\textbf{Answer verification.}}
Lastly, annotators are asked to verify the correctness of the model's final prediction. Since large language models tend to output incorrect conclusions when the explanation is factually mistaken~\citep{wang2022self}, it is very likely that the answer would be wrong when the original explanation got a low score in S3.
\section{Analysis of Explanation Data}
In this section, we analyze what error cases are abundant within an explanation and show how they can be revised using evidence documents retrieved by \textbf{CoTEVer}. As mentioned in Section~\ref{section3.1:prompting}, we adopt a Self-Ask style prompt and use \textsc{text-davinci-002}~\citep{ouyang2022training} to generate a corresponding explanation and answer for the train set of StrategyQA~\citep{geva2021did}. Then, we sample 300 instances where the prediction is incorrect, ask annotators to classify the error type and revise the explanation using \textbf{CoTEVer}. 

While we analyze the error types of explanations using human evaluation, automatic evaluation metrics proposed to measure the quality of a given explanation~\citep{golovneva2022roscoe, chen2022rev} is another promising direction, and we leave for future work. Also, we provide more detail of the human evaluation experiment process in Appendix~\ref{section:experimentdetail}. Table~\ref{tab:update} shows three frequently observed errors types, \textbf{Insufficient Knowledge}, \textbf{Out of Date} and \textbf{Wrong Fact} along with the corresponding percentage among the error cases (18\%, 22\%, 43\% respectively).\\
\newline
\noindent{\textbf{Insufficient Knowledge.}} 
It is well known that language models mainly learn from high-frequency
patterns and largely fail when tested on low resource tasks such as few-shot learning~\cite{tanzer2021bert}. Such behavior can be seen in the first example of Table~\ref{tab:update}. In general, it may be correct that crabs, lobsters and shrimp live in the oceans. However, the important point of the question is whether \textit{all} crustaceans live in the ocean, making the generated explanation  \textit{insufficient}. The knowledge needed in such situation is included in the evidence documents, where it indicates that crabs and shrimp also live in freshwater. \\

\noindent{\textbf{Out of Date.}}
The static nature of the text data that large language models are trained on makes it difficult to cope with rapidly changing real world situations~\cite{jang2021towards}. For instance, in the second example of Table~\ref{tab:update}, bitcoin is a highly volatile asset that has gone up significantly in the past few years. According to the retrieved evidence document, it is no longer \$6000 but actually more than \$20k which exceeds the price of a Volkswagen Jetta. These types of updates need to be done frequently through retrieval of up-to-date documents.\\

\noindent{\textbf{Wrong Fact.}} 
As shown in the third example of Table~\ref{tab:update}, large language models also generate false facts within the explanation. In this case, the first step within the explanation quoting, "The Hooke Sea is a sea in Antarctica." is not true. Because the Hooke Sea is not in Antarctica but on Mars, it isn't actually a sea, eliminating the lifeboat scenario. This fact can also be found in the retrieved document.

\section{How to Utilize Explanation Data gathered with \textbf{CoTEVer}} \label{sec:5}
In this section, we suggest three promising directions on how the explanation data collected with \textbf{CoTEVer} can be utilized.
We define $\mathcal{E}$ and $\mathcal{A}$ to be the original explanation and answer generated by a language model, respectively. Similarly, the revised explanation and answer from the annotator can be defined as $\mathcal{E}^{*}$ and $\mathcal{A}^{*}$. Explanations consist of pairs of sub-questions ${sq}_i$ and sub-answers ${sa}_i$ which brings the following definition:
\newline
\begin{itemize}
    \item Explanation $\mathcal{E}$ with $N$ pairs of $e_i=({sq}_i, {sa}_i)$ is $\mathcal{E}=\{e_i\}_{i=1}^{N}$
    \item A revised explanation $\mathcal{E}^{*}$ with $N^{*}$ pairs of $e^*=({sq^{*}}_i, {sa^{*}}_i)$ is  $\mathcal{E}^{*}=\{e^{*}_i\}_{i=1}^{N^{*}}$
\end{itemize}

Now for an explanation, sets of documents $\mathcal{D}_i$ are retrieved  for each pair $e_i$, based on ${sq}_i$. Within $\mathcal{D}_i$, we define the top-$k^{th}$ document aligned by the re-ranking module as $\mathcal{D}_{i}^k$.
Finally, $\Tilde{\mathcal{D}_i}$ is defined as the evidence document chosen by the annotator upon the set $\mathcal{D}_i$.


\subsection{Chain of Thought Fine-tuning}
\citet{chung2022scaling} indicated that fine-tuning language models to generate an explanation is effective to improve reasoning abilities of language models. We suggest training a model using the revised explanation collected by \textbf{CoTEVer} instead of using manually collected explanations. The objective could be formalized such as:
\begin{equation}\label{cot_loss_estar}
    \mbox{${\cal L}$}_{e^{*}} = - \sum_{i=1}^{|\mathcal{E^{*}}|} \sum_{j=1}^{|e^{*}_i|} \log P(e^{*}_{i,j}|e^{*}_{<i},e^{*}_{i,<j})
\end{equation}

\begin{equation}\label{cot_loss_astar}
    \mbox{${\cal L}$}_{a^{*}} = - \sum_{i=1}^{|\mathcal{A^{*}}|} \log P(a^{*}_{i}|a_{<i},\mathcal{E^{*}})
\end{equation}

where the $i^{th}$ explanation $e^{*}$ consists of $|e^{*}_i|$ tokens. Note that in CoT Fine-tuning, the explanation is first generated by conditioning on the question, and then the answer is generated by conditioning on the question and explanation (explain-and-generate).\\

\noindent{\textbf{Unlikelihood Training}} In addition to using the revised explanation to teach language models to generate an explanation coupled with the final prediction, we also suggest using the incorrect explanations for knowledge unlearning via unlikelihood training~\citep{welleck2019neural}. Prior work proposed that simply negating the original cross entropy loss is effective in knowledge unlearning~\citep{jang2022knowledge}. In the case of explanation data, models can forget incorrect explanations and learn the correct explanations instead. Using the rating score provided by the annotators, we could define how much room of improvement there was between the original explanation and the revised explanation. We could use `original explanations with relatively low scores' among the collected explanations as hard negatives. Then, the objective could be formalized such as:

\begin{equation}\label{cot_loss_e}
    \mbox{${\cal L}$}_{e} = - \sum_{i=1}^{|\mathcal{E}|} \sum_{j=1}^{|e_i|} \log (1  - P(e_{i,j}|e_{<i},e_{i,<j}))
\end{equation}

Future work could consider analyzing whether forgetting the incorrect explanation before learning the correct explanation is more effective, or vice versa. Also, a more sophisticated definition of how to determine `incorrect explanations' and `correct explanations' using the user's feedback score could be explored.

\subsection{Knowledge-Intensive Tasks}
As we show in Table \ref{tab:update}, large language models tend to generate unfaithful explanations, which is especially problematic when solving knowledge-intensive tasks~\citep{lewis2020retrieval}. We suggest two approaches that could resolve this issue by building datasets for fact verification and information retrieval from the revised explanations and the evidence documents.\\ 

\noindent{\textbf{Fact Verification.}} Following the task definition of FEVER~\citep{thorne2018fever}, we define labels for each pair of sub-answer ${sa}_i$ and a evidence document from $\mathcal{D}_i$ as either \textsc{Supported}, \textsc{Refuted}, and \textsc{NotEnoughInfo}. 

Since the annotators use $\Tilde{\mathcal{D}_{i}}$ as evidence when finding contradictions, ${sa}_i$ rated as 1 and $\Tilde{\mathcal{D}_i}$ can be labeled as \textsc{Refuted}. Similarly, the pair of ${sa}^{*}_{i}$\footnote{${sa}^{*}_{i}$ where the original ${sa}_i$ was rated as 1, which is the lowest score.} and document $\Tilde{\mathcal{D}_{i}}$ can be labeled as \textsc{Supported}. As low-ranked documents $\mathcal{D}_{i}^{10}$ from our re-ranking module are less likely to contain information that supports nor refutes the explanations, we use them as examples for \textsc{NotEnoughInfo}. The fact verification data obtained with \textbf{CoTEVer} could be used to to train a factual error correction model~\citep{thorne2021evidence}.\\



\noindent{\textbf{Information Retrieval.}} \citet{karpukhin2020dense} explains that using negative examples helps substantially, whilst they mitigated the difficulty in obtaining them via setting in-batch negatives. \textbf{CoTEVer} is effective to acquire hard negative as well as positive pairs using the sub-questions ${sq}_i$ and a evidence document from $\mathcal{D}_i$. 

Since the annotators find $\Tilde{\mathcal{D}_{i}}$ to contain the most helpful information when revising ${sa}_i$ rated as 1 to ${sa}^{*}_{i}$, $\Tilde{\mathcal{D}_{i}}$ would form a positive relation with ${sq}_i$. Meanwhile, $\mathcal{D}_{i}^{10}$, which was ranked low by our re-ranking module would serve as a hard negative for ${sq}_i$. The information retrieval data obtained with \textbf{CoTEVer} could be used to train a enhanced dense embedding model~\citep{gao2021simcse, chuang2022diffcse}.

\section{Conclusion}
In this work, we introduce \textbf{CoTEver}, an interactive annotation framework designed to verify unfaithful outputs and gather truthful explanation data from annotators. To reduce the cost of manually searching for evidence while verifying an explanation, we provide supporting evidence documents via a search engine. Next, we analyze some of the abundant reasons where large language models generated incorrect explanations. Also, we suggest three directions on how explanation data gathered with \textbf{CoTEVer} can be utilized. We hope \textbf{CoTEVer} will contribute to gather high quality explanation data used for future research.

\section*{Acknowledgements}
We thank Minkyeong Moon for helping make the demonstration video; Sangwon Park, Sehwan Jeon, Imsung Yu, and Donghwan Park for helping implement the frontend and backend of CoTEVer; Seonghyeon Ye, Hoyeon Chang, Joel Jang, Yongho Song, and anonymous reviewers for helpful feedback. This work was partly supported by Institute of Information \& communications Technology Planning \& Evaluation (IITP) grant funded by the Korea government(MSIT) (No. 2020-0-01361, Artificial Intelligence Graduate School Program (Yonsei University)), (No.2021-0-02068, Artificial Intelligence Innovation Hub), and (No. 2022-0-00077, AI Technology Development for Commonsense Extraction, Reasoning, and Inference from Heterogeneous Data). Jinyoung Yeo is the corresponding author.


\bibliography{eacl2023}

\begin{thebibliography}{49}
\expandafter\ifx\csname natexlab\endcsname\relax\def\natexlab#1{#1}\fi

\bibitem[{Bach et~al.(2022)Bach, Sanh, Yong, Webson, Raffel, Nayak, Sharma,
  Kim, Bari, Fevry et~al.}]{bach2022promptsource}
Stephen~H Bach, Victor Sanh, Zheng-Xin Yong, Albert Webson, Colin Raffel,
  Nihal~V Nayak, Abheesht Sharma, Taewoon Kim, M~Saiful Bari, Thibault Fevry,
  et~al. 2022.
\newblock Promptsource: An integrated development environment and repository
  for natural language prompts.
\newblock \emph{arXiv preprint arXiv:2202.01279}.

\bibitem[{Brown et~al.(2020)Brown, Mann, Ryder, Subbiah, Kaplan, Dhariwal,
  Neelakantan, Shyam, Sastry, Askell et~al.}]{brown2020language}
Tom Brown, Benjamin Mann, Nick Ryder, Melanie Subbiah, Jared~D Kaplan, Prafulla
  Dhariwal, Arvind Neelakantan, Pranav Shyam, Girish Sastry, Amanda Askell,
  et~al. 2020.
\newblock Language models are few-shot learners.
\newblock \emph{Advances in neural information processing systems},
  33:1877--1901.

\bibitem[{Chen et~al.(2022)Chen, Brahman, Ren, Ji, Choi, and
  Swayamdipta}]{chen2022rev}
Hanjie Chen, Faeze Brahman, Xiang Ren, Yangfeng Ji, Yejin Choi, and Swabha
  Swayamdipta. 2022.
\newblock Rev: Information-theoretic evaluation of free-text rationales.
\newblock \emph{arXiv preprint arXiv:2210.04982}.

\bibitem[{Chuang et~al.(2022)Chuang, Dangovski, Luo, Zhang, Chang,
  Solja{\v{c}}i{\'c}, Li, Yih, Kim, and Glass}]{chuang2022diffcse}
Yung-Sung Chuang, Rumen Dangovski, Hongyin Luo, Yang Zhang, Shiyu Chang, Marin
  Solja{\v{c}}i{\'c}, Shang-Wen Li, Wen-tau Yih, Yoon Kim, and James Glass.
  2022.
\newblock Diffcse: Difference-based contrastive learning for sentence
  embeddings.
\newblock \emph{arXiv preprint arXiv:2204.10298}.

\bibitem[{Chung et~al.(2022)Chung, Hou, Longpre, Zoph, Tay, Fedus, Li, Wang,
  Dehghani, Brahma et~al.}]{chung2022scaling}
Hyung~Won Chung, Le~Hou, Shayne Longpre, Barret Zoph, Yi~Tay, William Fedus,
  Eric Li, Xuezhi Wang, Mostafa Dehghani, Siddhartha Brahma, et~al. 2022.
\newblock Scaling instruction-finetuned language models.
\newblock \emph{arXiv preprint arXiv:2210.11416}.

\bibitem[{Cobbe et~al.(2021)Cobbe, Kosaraju, Bavarian, Hilton, Nakano, Hesse,
  and Schulman}]{cobbe2021training}
Karl Cobbe, Vineet Kosaraju, Mohammad Bavarian, Jacob Hilton, Reiichiro Nakano,
  Christopher Hesse, and John Schulman. 2021.
\newblock Training verifiers to solve math word problems.
\newblock \emph{arXiv preprint arXiv:2110.14168}.

\bibitem[{Du et~al.(2022)Du, Kim, Raheja, Kumar, and Kang}]{du2022read}
Wanyu Du, Zae~Myung Kim, Vipul Raheja, Dhruv Kumar, and Dongyeop Kang. 2022.
\newblock Read, revise, repeat: A system demonstration for human-in-the-loop
  iterative text revision.
\newblock In \emph{Proceedings of the First Workshop on Intelligent and
  Interactive Writing Assistants (In2Writing 2022)}, pages 96--108.

\bibitem[{Friedrich et~al.(2021)Friedrich, Gashteovski, Yu, Kotnis, Lawrence,
  Niepert, and Glava{\v{s}}}]{friedrich2021annie}
Niklas Friedrich, Kiril Gashteovski, Mingying Yu, Bhushan Kotnis, Carolin
  Lawrence, Mathias Niepert, and Goran Glava{\v{s}}. 2021.
\newblock Annie: An annotation platform for constructing complete open
  information extraction benchmark.
\newblock \emph{arXiv preprint arXiv:2109.07464}.

\bibitem[{Gao et~al.(2022)Gao, Dai, Pasupat, Chen, Chaganty, Fan, Zhao, Lao,
  Lee, Juan et~al.}]{gao2022attributed}
Luyu Gao, Zhuyun Dai, Panupong Pasupat, Anthony Chen, Arun~Tejasvi Chaganty,
  Yicheng Fan, Vincent~Y Zhao, Ni~Lao, Hongrae Lee, Da-Cheng Juan, et~al. 2022.
\newblock Attributed text generation via post-hoc research and revision.
\newblock \emph{arXiv preprint arXiv:2210.08726}.

\bibitem[{Gao et~al.(2021)Gao, Yao, and Chen}]{gao2021simcse}
Tianyu Gao, Xingcheng Yao, and Danqi Chen. 2021.
\newblock Simcse: Simple contrastive learning of sentence embeddings.
\newblock In \emph{Proceedings of the 2021 Conference on Empirical Methods in
  Natural Language Processing}, pages 6894--6910.

\bibitem[{Geva et~al.(2021)Geva, Khashabi, Segal, Khot, Roth, and
  Berant}]{geva2021did}
Mor Geva, Daniel Khashabi, Elad Segal, Tushar Khot, Dan Roth, and Jonathan
  Berant. 2021.
\newblock Did aristotle use a laptop? a question answering benchmark with
  implicit reasoning strategies.
\newblock \emph{Transactions of the Association for Computational Linguistics},
  9:346--361.

\bibitem[{Golovneva et~al.(2022)Golovneva, Chen, Poff, Corredor, Zettlemoyer,
  Fazel-Zarandi, and Celikyilmaz}]{golovneva2022roscoe}
Olga Golovneva, Moya Chen, Spencer Poff, Martin Corredor, Luke Zettlemoyer,
  Maryam Fazel-Zarandi, and Asli Celikyilmaz. 2022.
\newblock Roscoe: A suite of metrics for scoring step-by-step reasoning.
\newblock \emph{arXiv preprint arXiv:2212.07919}.

\bibitem[{G{\"o}tze et~al.(2022)G{\"o}tze, Paetzel-Pr{\"u}smann, Liermann,
  Diekmann, and Schlangen}]{gotze2022slurk}
Jana G{\"o}tze, Maike Paetzel-Pr{\"u}smann, Wencke Liermann, Tim Diekmann, and
  David Schlangen. 2022.
\newblock The slurk interaction server framework: Better data for better dialog
  models.
\newblock \emph{arXiv preprint arXiv:2202.01155}.

\bibitem[{Huang et~al.(2022)Huang, Gu, Hou, Wu, Wang, Yu, and
  Han}]{huang2022large}
Jiaxin Huang, Shixiang~Shane Gu, Le~Hou, Yuexin Wu, Xuezhi Wang, Hongkun Yu,
  and Jiawei Han. 2022.
\newblock Large language models can self-improve.
\newblock \emph{arXiv preprint arXiv:2210.11610}.

\bibitem[{Jang et~al.(2023)Jang, Kim, Ye, Kim, Logeswaran, Lee, Lee, and
  Seo}]{jang2023exploring}
Joel Jang, Seungone Kim, Seonghyeon Ye, Doyoung Kim, Lajanugen Logeswaran,
  Moontae Lee, Kyungjae Lee, and Minjoon Seo. 2023.
\newblock Exploring the benefits of training expert language models over
  instruction tuning.
\newblock \emph{arXiv preprint arXiv:2302.03202}.

\bibitem[{Jang et~al.(2021)Jang, Ye, Yang, Shin, Han, Kim, Choi, and
  Seo}]{jang2021towards}
Joel Jang, Seonghyeon Ye, Sohee Yang, Joongbo Shin, Janghoon Han, Gyeonghun
  Kim, Stanley~Jungkyu Choi, and Minjoon Seo. 2021.
\newblock Towards continual knowledge learning of language models.
\newblock \emph{arXiv preprint arXiv:2110.03215}.

\bibitem[{Jang et~al.(2022)Jang, Yoon, Yang, Cha, Lee, Logeswaran, and
  Seo}]{jang2022knowledge}
Joel Jang, Dongkeun Yoon, Sohee Yang, Sungmin Cha, Moontae Lee, Lajanugen
  Logeswaran, and Minjoon Seo. 2022.
\newblock Knowledge unlearning for mitigating privacy risks in language models.
\newblock \emph{arXiv preprint arXiv:2210.01504}.

\bibitem[{Karpukhin et~al.(2020)Karpukhin, Oguz, Min, Lewis, Wu, Edunov, Chen,
  and Yih}]{karpukhin2020dense}
Vladimir Karpukhin, Barlas Oguz, Sewon Min, Patrick Lewis, Ledell Wu, Sergey
  Edunov, Danqi Chen, and Wen-tau Yih. 2020.
\newblock Dense passage retrieval for open-domain question answering.
\newblock In \emph{Proceedings of the 2020 Conference on Empirical Methods in
  Natural Language Processing (EMNLP)}, pages 6769--6781.

\bibitem[{Kojima et~al.(2022)Kojima, Gu, Reid, Matsuo, and
  Iwasawa}]{kojima2022large}
Takeshi Kojima, Shixiang~Shane Gu, Machel Reid, Yutaka Matsuo, and Yusuke
  Iwasawa. 2022.
\newblock Large language models are zero-shot reasoners.
\newblock \emph{arXiv preprint arXiv:2205.11916}.

\bibitem[{Lewis et~al.(2020)Lewis, Perez, Piktus, Petroni, Karpukhin, Goyal,
  K{\"u}ttler, Lewis, Yih, Rockt{\"a}schel et~al.}]{lewis2020retrieval}
Patrick Lewis, Ethan Perez, Aleksandra Piktus, Fabio Petroni, Vladimir
  Karpukhin, Naman Goyal, Heinrich K{\"u}ttler, Mike Lewis, Wen-tau Yih, Tim
  Rockt{\"a}schel, et~al. 2020.
\newblock Retrieval-augmented generation for knowledge-intensive nlp tasks.
\newblock \emph{Advances in Neural Information Processing Systems},
  33:9459--9474.

\bibitem[{Li et~al.(2022)Li, Lin, Zhang, Fu, Chen, Lou, and
  Chen}]{li2022advance}
Yifei Li, Zeqi Lin, Shizhuo Zhang, Qiang Fu, Bei Chen, Jian-Guang Lou, and
  Weizhu Chen. 2022.
\newblock On the advance of making language models better reasoners.
\newblock \emph{arXiv preprint arXiv:2206.02336}.

\bibitem[{Lin et~al.(2022{\natexlab{a}})Lin, Hilton, and
  Evans}]{lin2022truthfulqa}
Stephanie Lin, Jacob Hilton, and Owain Evans. 2022{\natexlab{a}}.
\newblock Truthfulqa: Measuring how models mimic human falsehoods.
\newblock In \emph{Proceedings of the 60th Annual Meeting of the Association
  for Computational Linguistics (Volume 1: Long Papers)}, pages 3214--3252.

\bibitem[{Lin et~al.(2022{\natexlab{b}})Lin, Ruan, Liang, Cai, Du, and
  Wang}]{lin2022dotat}
Yupian Lin, Tong Ruan, Ming Liang, Tingting Cai, Wen Du, and Yi~Wang.
  2022{\natexlab{b}}.
\newblock Dotat: A domain-oriented text annotation tool.
\newblock In \emph{Proceedings of the 60th Annual Meeting of the Association
  for Computational Linguistics: System Demonstrations}, pages 1--8.

\bibitem[{Ling et~al.(2017)Ling, Yogatama, Dyer, and Blunsom}]{ling2017program}
Wang Ling, Dani Yogatama, Chris Dyer, and Phil Blunsom. 2017.
\newblock Program induction by rationale generation: Learning to solve and
  explain algebraic word problems.
\newblock In \emph{Proceedings of the 55th Annual Meeting of the Association
  for Computational Linguistics (Volume 1: Long Papers)}, pages 158--167.

\bibitem[{Liu et~al.(2021)Liu, Shen, Zhang, Dolan, Carin, and
  Chen}]{liu2021makes}
Jiachang Liu, Dinghan Shen, Yizhe Zhang, Bill Dolan, Lawrence Carin, and Weizhu
  Chen. 2021.
\newblock What makes good in-context examples for gpt-$3 $?
\newblock \emph{arXiv preprint arXiv:2101.06804}.

\bibitem[{Min et~al.(2021)Min, Lewis, Zettlemoyer, and
  Hajishirzi}]{min2021metaicl}
Sewon Min, Mike Lewis, Luke Zettlemoyer, and Hannaneh Hajishirzi. 2021.
\newblock Metaicl: Learning to learn in context.
\newblock \emph{arXiv preprint arXiv:2110.15943}.

\bibitem[{Ni et~al.(2021)Ni, {\'A}brego, Constant, Ma, Hall, Cer, and
  Yang}]{ni2021sentence}
Jianmo Ni, Gustavo~Hern{\'a}ndez {\'A}brego, Noah Constant, Ji~Ma, Keith~B
  Hall, Daniel Cer, and Yinfei Yang. 2021.
\newblock Sentence-t5: Scalable sentence encoders from pre-trained text-to-text
  models.
\newblock \emph{arXiv preprint arXiv:2108.08877}.

\bibitem[{Ouyang et~al.(2022)Ouyang, Wu, Jiang, Almeida, Wainwright, Mishkin,
  Zhang, Agarwal, Slama, Ray et~al.}]{ouyang2022training}
Long Ouyang, Jeff Wu, Xu~Jiang, Diogo Almeida, Carroll~L Wainwright, Pamela
  Mishkin, Chong Zhang, Sandhini Agarwal, Katarina Slama, Alex Ray, et~al.
  2022.
\newblock Training language models to follow instructions with human feedback.
\newblock \emph{arXiv preprint arXiv:2203.02155}.

\bibitem[{Press et~al.(2022)Press, Zhang, Min, Schmidt, Smith, and
  Lewis}]{press2022measuring}
Ofir Press, Muru Zhang, Sewon Min, Ludwig Schmidt, Noah~A Smith, and Mike
  Lewis. 2022.
\newblock Measuring and narrowing the compositionality gap in language models.
\newblock \emph{arXiv preprint arXiv:2210.03350}.

\bibitem[{Sanh et~al.(2021)Sanh, Webson, Raffel, Bach, Sutawika, Alyafeai,
  Chaffin, Stiegler, Scao, Raja et~al.}]{sanh2021multitask}
Victor Sanh, Albert Webson, Colin Raffel, Stephen~H Bach, Lintang Sutawika,
  Zaid Alyafeai, Antoine Chaffin, Arnaud Stiegler, Teven~Le Scao, Arun Raja,
  et~al. 2021.
\newblock Multitask prompted training enables zero-shot task generalization.
\newblock \emph{arXiv preprint arXiv:2110.08207}.

\bibitem[{Shen et~al.(2021)Shen, Yin, Li, Shang, Jiang, Zhang, and
  Liu}]{shen2021generate}
Jianhao Shen, Yichun Yin, Lin Li, Lifeng Shang, Xin Jiang, Ming Zhang, and Qun
  Liu. 2021.
\newblock Generate \& rank: A multi-task framework for math word problems.
\newblock \emph{arXiv preprint arXiv:2109.03034}.

\bibitem[{Shuster et~al.(2021)Shuster, Poff, Chen, Kiela, and
  Weston}]{shuster2021retrieval}
Kurt Shuster, Spencer Poff, Moya Chen, Douwe Kiela, and Jason Weston. 2021.
\newblock Retrieval augmentation reduces hallucination in conversation.
\newblock In \emph{Findings of the Association for Computational Linguistics:
  EMNLP 2021}, pages 3784--3803.

\bibitem[{Srivastava et~al.(2022)Srivastava, Rastogi, Rao, Shoeb, Abid, Fisch,
  Brown, Santoro, Gupta, Garriga-Alonso et~al.}]{srivastava2022beyond}
Aarohi Srivastava, Abhinav Rastogi, Abhishek Rao, Abu Awal~Md Shoeb, Abubakar
  Abid, Adam Fisch, Adam~R Brown, Adam Santoro, Aditya Gupta, Adri{\`a}
  Garriga-Alonso, et~al. 2022.
\newblock Beyond the imitation game: Quantifying and extrapolating the
  capabilities of language models.
\newblock \emph{arXiv preprint arXiv:2206.04615}.

\bibitem[{T{\"a}nzer et~al.(2021)T{\"a}nzer, Ruder, and Rei}]{tanzer2021bert}
Michael T{\"a}nzer, Sebastian Ruder, and Marek Rei. 2021.
\newblock Bert memorisation and pitfalls in low-resource scenarios.
\newblock \emph{arXiv preprint arXiv:2105.00828}.

\bibitem[{Thoppilan et~al.(2022)Thoppilan, De~Freitas, Hall, Shazeer,
  Kulshreshtha, Cheng, Jin, Bos, Baker, Du et~al.}]{thoppilan2022lamda}
Romal Thoppilan, Daniel De~Freitas, Jamie Hall, Noam Shazeer, Apoorv
  Kulshreshtha, Heng-Tze Cheng, Alicia Jin, Taylor Bos, Leslie Baker, Yu~Du,
  et~al. 2022.
\newblock Lamda: Language models for dialog applications.
\newblock \emph{arXiv preprint arXiv:2201.08239}.

\bibitem[{Thorne and Vlachos(2021)}]{thorne2021evidence}
James Thorne and Andreas Vlachos. 2021.
\newblock Evidence-based factual error correction.
\newblock In \emph{Proceedings of the 59th Annual Meeting of the Association
  for Computational Linguistics and the 11th International Joint Conference on
  Natural Language Processing (Volume 1: Long Papers)}, pages 3298--3309.

\bibitem[{Thorne et~al.(2018)Thorne, Vlachos, Christodoulopoulos, and
  Mittal}]{thorne2018fever}
James Thorne, Andreas Vlachos, Christos Christodoulopoulos, and Arpit Mittal.
  2018.
\newblock Fever: a large-scale dataset for fact extraction and verification.
\newblock In \emph{Proceedings of the 2018 Conference of the North American
  Chapter of the Association for Computational Linguistics: Human Language
  Technologies, Volume 1 (Long Papers)}, pages 809--819.

\bibitem[{Thrush et~al.(2022)Thrush, Tirumala, Gupta, Bartolo, Rodriguez, Kane,
  Rojas, Mattson, Williams, and Kiela}]{thrush2022dynatask}
Tristan Thrush, Kushal Tirumala, Anmol Gupta, Max Bartolo, Pedro Rodriguez,
  Tariq Kane, William~Gaviria Rojas, Peter Mattson, Adina Williams, and Douwe
  Kiela. 2022.
\newblock Dynatask: A framework for creating dynamic ai benchmark tasks.
\newblock \emph{arXiv preprint arXiv:2204.01906}.

\bibitem[{Wallace et~al.(2019)Wallace, Rodriguez, Feng, Yamada, and
  Boyd-Graber}]{wallace2019trick}
Eric Wallace, Pedro Rodriguez, Shi Feng, Ikuya Yamada, and Jordan Boyd-Graber.
  2019.
\newblock Trick me if you can: Human-in-the-loop generation of adversarial
  examples for question answering.
\newblock \emph{Transactions of the Association for Computational Linguistics},
  7:387--401.

\bibitem[{Wang et~al.(2022)Wang, Wei, Schuurmans, Le, Chi, and
  Zhou}]{wang2022self}
Xuezhi Wang, Jason Wei, Dale Schuurmans, Quoc Le, Ed~Chi, and Denny Zhou. 2022.
\newblock Self-consistency improves chain of thought reasoning in language
  models.
\newblock \emph{arXiv preprint arXiv:2203.11171}.

\bibitem[{Weber et~al.(2021)Weber, Piovano, and Bradford}]{weber2021better}
Verena Weber, Enrico Piovano, and Melanie Bradford. 2021.
\newblock It is better to verify: Semi-supervised learning with a human in the
  loop for large-scale nlu models.
\newblock In \emph{Proceedings of the Second Workshop on Data Science with
  Human in the Loop: Language Advances}, pages 8--15.

\bibitem[{Wei et~al.(2021)Wei, Bosma, Zhao, Guu, Yu, Lester, Du, Dai, and
  Le}]{wei2021finetuned}
Jason Wei, Maarten Bosma, Vincent~Y Zhao, Kelvin Guu, Adams~Wei Yu, Brian
  Lester, Nan Du, Andrew~M Dai, and Quoc~V Le. 2021.
\newblock Finetuned language models are zero-shot learners.
\newblock \emph{arXiv preprint arXiv:2109.01652}.

\bibitem[{Wei et~al.(2022{\natexlab{a}})Wei, Tay, Bommasani, Raffel, Zoph,
  Borgeaud, Yogatama, Bosma, Zhou, Metzler et~al.}]{wei2022emergent}
Jason Wei, Yi~Tay, Rishi Bommasani, Colin Raffel, Barret Zoph, Sebastian
  Borgeaud, Dani Yogatama, Maarten Bosma, Denny Zhou, Donald Metzler, et~al.
  2022{\natexlab{a}}.
\newblock Emergent abilities of large language models.
\newblock \emph{arXiv preprint arXiv:2206.07682}.

\bibitem[{Wei et~al.(2022{\natexlab{b}})Wei, Wang, Schuurmans, Bosma, Chi, Le,
  and Zhou}]{wei2022chain}
Jason Wei, Xuezhi Wang, Dale Schuurmans, Maarten Bosma, Ed~Chi, Quoc Le, and
  Denny Zhou. 2022{\natexlab{b}}.
\newblock Chain of thought prompting elicits reasoning in large language
  models.
\newblock \emph{arXiv preprint arXiv:2201.11903}.

\bibitem[{Welleck et~al.(2019)Welleck, Kulikov, Roller, Dinan, Cho, and
  Weston}]{welleck2019neural}
Sean Welleck, Ilia Kulikov, Stephen Roller, Emily Dinan, Kyunghyun Cho, and
  Jason Weston. 2019.
\newblock Neural text generation with unlikelihood training.
\newblock \emph{arXiv preprint arXiv:1908.04319}.

\bibitem[{West et~al.(2021)West, Bhagavatula, Hessel, Hwang, Jiang, Bras, Lu,
  Welleck, and Choi}]{west2021symbolic}
Peter West, Chandra Bhagavatula, Jack Hessel, Jena~D Hwang, Liwei Jiang,
  Ronan~Le Bras, Ximing Lu, Sean Welleck, and Yejin Choi. 2021.
\newblock Symbolic knowledge distillation: from general language models to
  commonsense models.
\newblock \emph{arXiv preprint arXiv:2110.07178}.

\bibitem[{Wiegreffe and Marasovic(2021)}]{wiegreffe2021teach}
Sarah Wiegreffe and Ana Marasovic. 2021.
\newblock Teach me to explain: A review of datasets for explainable natural
  language processing.
\newblock In \emph{Thirty-fifth Conference on Neural Information Processing
  Systems Datasets and Benchmarks Track (Round 1)}.

\bibitem[{Ye et~al.(2022)Ye, Kim, Jang, Shin, and Seo}]{ye2022guess}
Seonghyeon Ye, Doyoung Kim, Joel Jang, Joongbo Shin, and Minjoon Seo. 2022.
\newblock Guess the instruction! making language models stronger zero-shot
  learners.
\newblock \emph{arXiv preprint arXiv:2210.02969}.

\bibitem[{Zelikman et~al.(2022)Zelikman, Wu, and Goodman}]{zelikman2022star}
Eric Zelikman, Yuhuai Wu, and Noah~D Goodman. 2022.
\newblock Star: Bootstrapping reasoning with reasoning.
\newblock \emph{arXiv preprint arXiv:2203.14465}.

\end{thebibliography}
\bibliographystyle{acl_natbib}

\appendix
\section{Link to Video \& Code}
The link to our video and code is as follows:
\begin{enumerate}
    \item \textbf{Demonstration Video}: \href{https://www.youtube.com/watch?v=IKT6dVxp_qE}{Link}
    \item \textbf{Official Code}: \href{https://github.com/SeungoneKim/CoTEVer}{Link}
\end{enumerate}

\section{Experiment Details for Human Evaluation}\label{section:experimentdetail}
Following \citet{wei2022chain}, we use the open-domain setting (question-only set) of StrategyQA~\citep{geva2021did} from \citet{srivastava2022beyond}. We use \textsc{text-davinci-002} to generate explanations. We set the temperature as 0. 

The 6-shot prompt we used are shown in Table~\ref{table:strategy_qa_prompt}. Our prompt are divided into sub-questions and sub-answers where the sub-questions are used as a query for retrieving the evidence documents.

\begin{table}[H]
\resizebox{\columnwidth}{!}{%
\begin{tabular}{|cc|}
\toprule
\multicolumn{2}{|c|}{\textbf{strategyQA}}                                             \\ \midrule
\multicolumn{1}{|c|}{CoT~\citep{wei2022chain}} & CoTEVer (Ours) \\ \midrule
\multicolumn{1}{|c|}{65.4}                       & 70.52                     \\ \midrule
\end{tabular}%
}
\caption{Few-shot Prompting accuracy on StrategyQA(question-only set). Our prompt consists of sub-questions and sub-answers.}
\label{tab:strategyqa}
\end{table}

Table~\ref{tab:strategyqa} shows the performance when using our designed prompt. Although our purpose of consisting prompts with sub-questions was for evidence retrieval, Self-Ask~\citep{press2022measuring} style prompts achieves better performance compared to the prompts used in \citet{wei2022chain}. Also, these results support the fact that the incorrect explanations(29.18\%) were not generated due to the quality of our prompt. 

To analyze the error types, we sample 300 incorrect instances since the explanation is likely to be wrong when the prediction is incorrect~\citep{wang2022self}. We ask 20 annotators with background in deep learning and proficient English skills to 1)classify the error type and 2)revise the explanation using \textbf{CoTEVer}. While the error types introduced in Table~\ref{tab:update} (total 83\%) could be revised based on the supporting evidence documents, 17\% were error types were GPT-3 didn't generate a final prediction by keep repeating itself, or error types where the knowledge was all correct, but the reasoning was incorrect. In this case, retrieving evidence documents does not help during explanation verification.

\section{Limitations}
The following are the limitations of \textbf{CoTEVer}.\\

\noindent{\textbf{Dependency on Prompt Design.}} While we experimented with prompts from StrategyQA, where only yes/no questions are given, annotators could type in any form of questions such as multiple choice questions or `why' style questions. In this case, the prompts used to query a large language model should be changed. Therefore, we provide functionality where the researcher using \textbf{CoTEVer} could change the prompt. To prevent unexpected behavior of large language models, it is recommended to instruct annotators what types of questions they should ask. Also beside output format, large language models are also sensitive to the domain of the question. For example, when annotators ask questions requiring financial of legal domain knowledge, the prompts we provided might not perform well. In this case, researchers should change the prompts to their needs.\\

\noindent{\textbf{Prevention on Usage for Unethical Use Cases.}} Large language models often tend to generate toxic outputs. This is a critical issue when annotators use our framework. Our annotation toolkit is built on top of the Open AI API, where there has been various attempts to mitigate the toxic behavior of GPT-3. For instance, their \textsc{text-davinci-002} version uses reinforcement learning to align towards non-toxic language.\\

\noindent{\textbf{Cases when Explanation is wrong due to In-correct Reasoning.}} While \textbf{CoTEVer} could supplement insufficient knowledge for annotators, it couldn't tell which step within the explanation contains incorrect reasoning. 

\begin{table*}[t!]
\begin{center}\small
\setlength{\extrarowheight}{2pt}
{\begin{tabular}{|l|l|l|} 
\toprule
Error  Type & \multicolumn{2}{l|}{Example} \\ 
\midrule
\multirow{14}{*}{\begin{tabular}[c]{@{}l@{}}Incomplete \\Answer\end{tabular}} & \multicolumn{2}{l|}{Is Black Lives Matter connected with capsaicin?}                                                                             \\ 
\cline{2-3}
& \#Q1 : What is capsaicin?                                    & \multirow{13}{*}{\begin{tabular}[c]{@{}l@{}}
Capsaicin (8-methyl-N-vanillyl-6-none
\\namide) is an active component of\\
chili peppers, which are plants \\
belonging to the genus Capsicum. \\
It is a chemical irritant for mammals,\\
including humans, and produces a\\
sensation of burning in any tissue\\
with which it comes into contact.\end{tabular}}  \\
& \#A1 : Capsaicin is a chemical compound                      &  \\
& that can be found in chili peppers. & \\
& \#Q1 : What is Black Lives Matter?                     & \\
& \#A1 : Black Lives Matter is a movement that fights against  &  \\
& racism and violence towards black people. & \\
& \#Q2 : Is Black Lives Matter connected with capsaicin?       & \\
& \#A2 : Since Black Lives Matter is a movement that fights &  \\
& against racism and violence towards black people,            & \\
& and capsaicin is a chemical compound that can be found  &  \\
& in chili peppers, Black Lives Matter is not connected        &  \\
& with capsaicin.                                              & \\
& So the answer is no.                                         & \\ 
\midrule
\multirow{9}{*}{Wrong Fact}                                                   & \multicolumn{2}{l|}{Does Soylent use Pea for their source of protein?}  \\ 
\cline{2-3}& \#Q1 : What is Soylent?  & \multirow{8}{*}{\begin{tabular}[c]{@{}l@{}}
Soy protein isolate is the ingredient\\
we use in Soylent as our source \\
of protein.\end{tabular}} \\
& \#A1 : Soylent is a meal replacement drink.&  \\
& \#Q2 : What is the source of protein for Soylent?& \\
& \#A2 : Soylent uses pea protein.                             &   \\
& \#Q3 : Does Soylent use Pea for their source                 &  \\
& \#A3 : Since Soylent uses pea protein, Soylent uses pea for  &  \\
& their source of protein.                                     &  \\
& So the answer is yes.                                        & \\ 
\midrule
\multirow{13}{*}{\begin{tabular}[c]{@{}l@{}}Update\\Required\end{tabular}}    & \multicolumn{2}{l|}{Has Kelly Clarkson outsold season 4 American Idol winner?} \\ 
\cline{2-3}
& \#Q1 : Who won season 4 of American Idol?                    & \multirow{6}{*}{\begin{tabular}[c]{@{}l@{}}
To date, Clarkson has sold over \\
28 million albums and 54 million \\
singles worldwide.\end{tabular}}                   \\
& \#A1 : Carrie Underwood won season 4 of American Idol.       &   \\
 & \#Q2 : How many albums did Carrie Underwood sell?            &   \\
& \#A2 : Carrie Underwood sold over 15 million albums.         &  \\
& \#Q2 : How many albums did Kelly Clarkson sell?              &  \\
& \#A2 : Kelly Clarkson sold over 25 million albums.           &  \\ 
\cline{3-3}
& \#Q3 : Has Kelly Clarkson outsold season 4 American Idol     & \multirow{6}{*}{\begin{tabular}[c]{@{}l@{}}
Throughout her career, Underwood has \\
sold 66 million albums worldwide. In\\
the United States, she remains the best\\
-selling album artist from American \\
Idol franchise with a total of 16.3\\
million albums sold\\
\end{tabular}}  \\
& winner?                                                      &                                                            \\
 & \#A3 : Since Kelly Clarkson sold over 25 million albums, and & \\
& Carrie Underwood sold over 15 million albums, Kelly     &  \\
& Clarkson sold more albums than Carrie Underwood.         &  \\
& So the answer is yes.  &  \\
\bottomrule
\end{tabular}}               
\end{center}
\caption{Additional Examples of frequent error types within explanations. The left side is the original explanation generated by
a language model, and the right side is the retrieved evidence document. The evidence documents could provide the
required knowledge when revising the original explanation.}
\label{tab:appendix}
\end{table*}

\begin{table}
\centering\small

\end{table}


\begin{figure*}[t!]
\centering
    \includegraphics[width=0.95\linewidth]{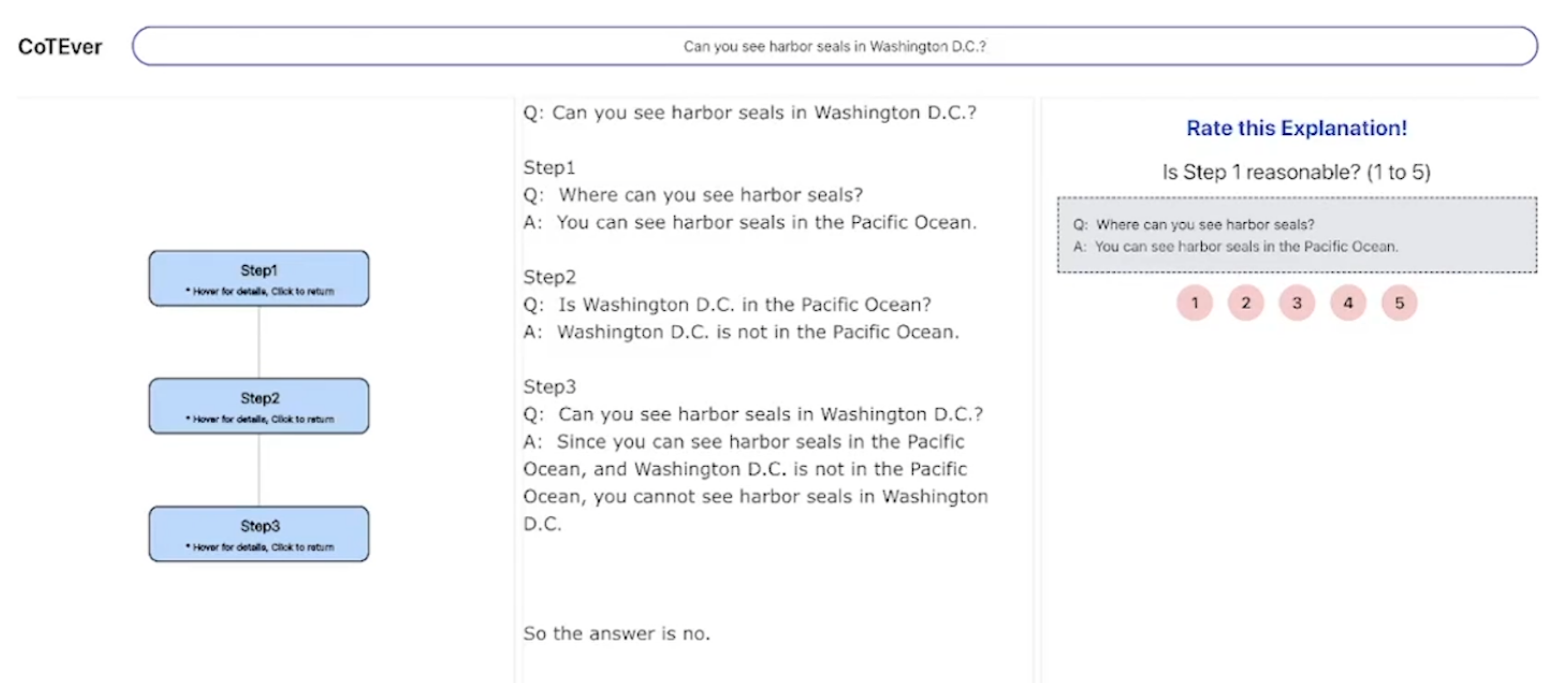}
    \caption{Snapshot of \textbf{CoTEVer}. Annotator gets to type in a question, and receive the output of a large language model(e.g., GPT-3).}
    \label{fig:cotever1}
\end{figure*}

\begin{figure*}[t!]
\centering
    \includegraphics[width=0.95\linewidth]{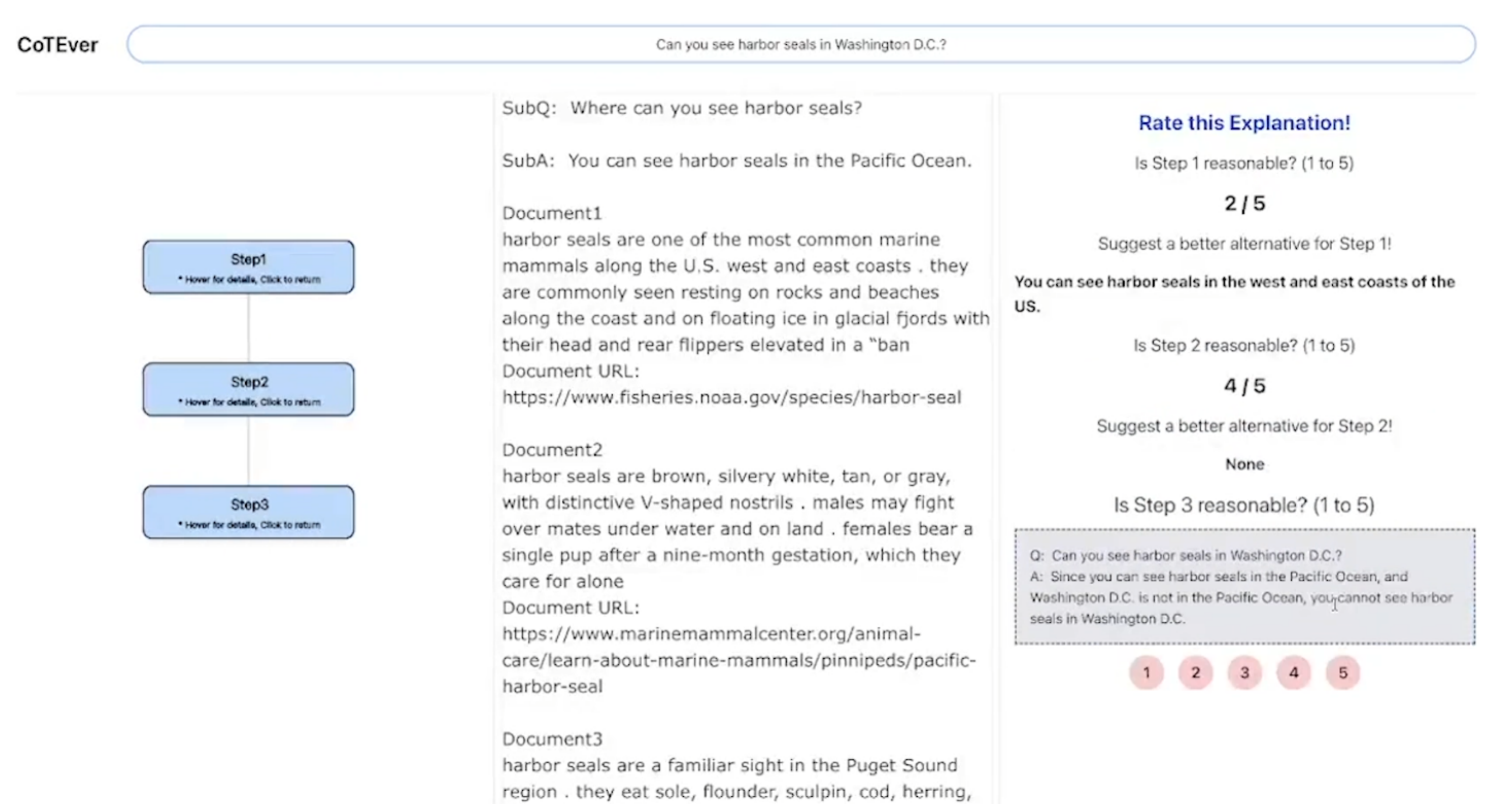}
    \caption{Snapshot of \textbf{CoTEVer}. Annotator could check the retrieved evidence documents in order to verify each step within the explanation.}
    \label{fig:cotever2}
\end{figure*}

\onecolumn
\begin{table}
\centering
\begin{tabular}{|l|} 
\toprule
Prompt \\ 
\midrule
\begin{tabular}[c]{@{}l@{}}{[}Example 1]\\Question: Do hamsters provide food for any animals? \\Output:\\Sub Question \#0 : What type of animals are hamsters? \\Sub Answer \#0 : Hamsters are prey animals. \\Sub Question \#1 : Can prey animals be food for other animals? \\Sub Answer \#1 : Prey are food for predators. \\Sub Question \#2 : Do hamsters provide food for any animals? \\Sub Answer \#2 : Since hamsters are prey animals, and prey are food \\for predators, hamsters provide food for some animals. \\Final Answer : So the answer is yes. \\\\{[}Example 2]\\Question: Could Brooke Shields succeed at University of Pennsylvania? \\Output:\\Sub question \#0 : What university did Brooke Shields went to? \\Sub answer \#0 : Brooke Shields went to Princeton University. \\Sub question \#1 : Did Brooke Shields succeed at Princeton University? \\Sub answer \#1 : At Princeton University, she got all As and Bs while \\pursing her bachelor's degree in French literature, meaning she had a \\successful school life. \\Sub question \#2 : How rigorous is Princeton University compared to \\University of Pennsylvania? \\Sub answer \#2 : Princeton University is about as academically rigorous \\as the University of Pennsylvania because they have a similar ranking\\according to U.S. News Rankings. \\Sub question \#3 : Could Brooke Shields succeed at University of Pennsylvania? \\Sub answer \#3 : Since University of Pennsylvania and University of \\Princeton are in similar circumstances, Brooke Shields has been successful \\in University of Princeton, Brooke Shields could also succeed at the \\University of Pennsylvania. \\So the answer is yes.\\\\{[}Example 3]\\Question: Hydrogen\textbackslash{}u2019s atomic number squared exceeds number of Spice Girls? \\Output:\\Sub question \#0 : What is the atomic number of Hydrogen? \\Sub answer \#0 : Hydrogen has an atomic number of 1. \\Sub question \#1 : What is 1 squared? \\Sub answer \#1 : 1 squared is 1. \\Sub question \#2 : How much Spice Girls are there? \\Sub answer \#2 : There are 5 Spice Girls. \\Sub question \#3 : Hydrogen\textbackslash{}u2019s atomic number squared exceeds number \\of Spice Girls? \\Sub answer \#3 : Since Hydrogen's atomic number squared is 1, the number of \\Spice Girls are 5, and 1 is smaller than 5, Hydrogen\textbackslash{}u2019s atomic number \\squared is less than the number of Spice Girls. \\So the answer is no.\end{tabular} \\
\bottomrule
\end{tabular}
\end{table}

\begin{table}
\centering
\begin{tabular}{|l|} 
\toprule
\begin{tabular}[c]{@{}l@{}}{[}Example 4]\\Question: Is it common to see frost during some college commencements?\\Output:\\Sub question \#0 : When does College commencement ceremonies usually happen? \\Sub answer \#0 : College commencement ceremonies can happen in December, \\May, and June. \\Sub question \#1 : Does it usually frost in December? \\Sub answer \#1 : December is in the winter, so there can be frost.\\Sub question \#2 : Is it common to see frost during some college commencements?\\Sub answer \#2 : Since there can be frost in December and a college \\commencement are held in December, there could be frost at some commencements.\\So the answer is yes.\\\\{[}Example 5]\\Question: Could a llama birth twice during War in Vietnam (1945-46)?\\Output:\\Sub question \#0 : How long was the Vietnam war?\\Sub answer \#0 : The War in Vietnam was 6 months.\\Sub question \#1 : How long is the gestation period?\\Sub answer \#1 : The gestation period for a llama is 11 months.\\Sub question \#2 : How long does it take for a llama to birth twice?\\Sub answer \#2 : Since the gestation period for a llama is 11 months, \\and 11 times 2 is 22, it will take 22 months.\\Sub question \#3 : Could a llama birth twice during War in Vietnam (1945-46)?\\Sub answer \#3 : Since it takes 22 months for a llama to birth twice, \\War in Vietnam was 6 months, and 22 is bigger than 6, llama could not \\give birth twice during the War in Vietnam.\\So the answer is no.\\\\{[}Example 6]\\Question: Would a pear sink in water?\\Output:\\Sub question \#0 : What is the density of a pear?\\Sub answer \#0 : The density of a pear is about 0.6g/cm3.\\Sub question \#1 : What is the density of water?\\Sub answer \#1 : The density of water is 1g/cm3.\\Sub question \#2 : Is the density of pear smaller than water?\\Sub answer \#2 : Since 0.6 is smaller than 1, the density of pear \\is smaller than water.\\Sub question \#3 : If the density of an object is less than water, what happens?\\Sub answer \#3 : Objects less dense than water float.\\Sub question \#4 : Would a pear sink in water?\\Sub answer \#4 : Since a pear has a smaller density than water, a pear would\\float.\\So the answer is no.\\\\{[}Example 7]\end{tabular} \\
\bottomrule
\end{tabular}
\caption{Prompt used to gather explanations for human evaluation experiments.}\label{table:strategy_qa_prompt}
\end{table}

\end{document}